\documentclass{article}

 \PassOptionsToPackage{numbers, compress}{natbib}

 \usepackage[preprint]{neurips_2019}

\usepackage[utf8]{inputenc} 
\usepackage[T1]{fontenc}    
\usepackage{hyperref}       
\usepackage{url}            
\usepackage{booktabs}       
\usepackage{amsfonts}       
\usepackage{nicefrac}       
\usepackage{microtype}      

\usepackage[bottom]{footmisc}

\usepackage{graphicx} 
\usepackage{subfigure}
\usepackage{color}
\usepackage{listings}
\usepackage{soul}
\usepackage{algorithm}
\usepackage[noend]{algpseudocode}
\usepackage{times}

\usepackage{amssymb}
\usepackage{amsmath,bm}
\usepackage{xcolor}
\usepackage{listings}

\usepackage{mathtools}

\usepackage{wrapfig}
\usepackage{dblfloatfix}
\usepackage{hyperref}
\usepackage{float}
\usepackage{multirow}

\makeatletter
\renewcommand\footnotesize{%
   \@setfontsize\footnotesize\@ixpt{11}%
   \abovedisplayskip 8\p@ \@plus2\p@ \@minus4\p@
   \abovedisplayshortskip \z@ \@plus\p@
   \belowdisplayshortskip 4\p@ \@plus2\p@ \@minus2\p@
   \def\@listi{\leftmargin\leftmargini
               \topsep 4\p@ \@plus2\p@ \@minus2\p@
               \parsep 2\p@ \@plus\p@ \@minus\p@
               \itemsep \parsep}%
   \belowdisplayskip \abovedisplayskip
}
\makeatother

\algnewcommand\algorithmicon{\textbf{Each}}
\algnewcommand\algorithmicond  {\textbf{Process}}
\algnewcommand\algorithmicondd  {\textbf{All processes of }}

\algnewcommand\algorithmicfrom{\textbf{}}
\algnewcommand\algorithmicperform{\textbf{does:}}
\algnewcommand\algorithmicperformd{\textbf{does:}}
\algnewcommand\algorithmicperformdd{\textbf{collectively do:}}
\algnewcommand{\LineComment}[1]{\State \(\blacktriangleright\) {\it #1}}
\algnewcommand\algorithmicforeach{\textbf{for each}}
\algdef{S}[FOR]{ForEach}[1]{\algorithmicforeach\ #1\ \algorithmicdo}

\algblockdefx[ON]{On}{EndOn}[2]
  {\algorithmicon\ #1\ \algorithmicfrom\ #2\ \algorithmicperform}
  {\algorithmicend\ \algorithmicon}
\algblockdefx[OND]{Ond}{EndOnd}[1]
  {\algorithmicond\ #1 \algorithmicperformd}
  {\algorithmicend\ \algorithmicon}
\algblockdefx[ONDD]{Ondd}{EndOndd}[1]
  {\algorithmicondd\ #1 \algorithmicperformdd}
  {\algorithmicend\ \algorithmicon}

\makeatletter
\ifthenelse{\equal{\ALG@noend}{t}}%
  {\algtext*{EndOn}}
  {}%
\ifthenelse{\equal{\ALG@noend}{t}}%
  {\algtext*{EndOndd}}
  {}%
\makeatother

\makeatletter
\ifthenelse{\equal{\ALG@noend}{t}}%
  {\algtext*{EndOnd}}
  {}%
\makeatother

\def\eqref#1{Eq. (\ref{#1})}
\def\figref#1{Fig. (\ref{#1})}
\def\secref#1{Sec. \ref{#1}}

\title{\Large Integrating Markov processes with structural causal modeling enables counterfactual inference in complex systems}

\author{%
  Robert Osazuwa Ness\\
  Gamalon Inc.\\
  robert.ness@gamalon.com\\
  \And
  Kaushal Paneri\\
  Northeastern University\\
  kaushalpaneri@gmail.com\\
  \And
  Olga Vitek\\
  Northeastern University\\
  o.vitek@northeastern.edu\\
}

\begin{document}

\vspace{-0.5cm}
\maketitle

\vspace{-0.5cm}
\begin{abstract}

This manuscript contributes a general and practical framework for casting a Markov process model of a system at equilibrium as a structural causal model, and carrying out counterfactual inference. Markov processes mathematically describe the mechanisms in the system, and predict the system's equilibrium behavior upon intervention, but do not support counterfactual inference. In contrast, structural causal models support counterfactual inference, but do not identify the mechanisms. This manuscript leverages the benefits of both approaches. We define the structural causal models in terms of the parameters and the equilibrium dynamics of the Markov process models, and counterfactual inference flows from these settings. The proposed approach alleviates the identifiability drawback of the structural causal models, in that the counterfactual inference is consistent with the counterfactual trajectories simulated from the Markov process model. We showcase the benefits of this framework in case studies of complex biomolecular systems with nonlinear dynamics. We illustrate that, in presence of Markov process model misspecification, counterfactual inference leverages prior data, and therefore estimates the outcome of an intervention more accurately than a direct simulation.

\end{abstract}

\vspace{-0.5cm}
\section{Introduction}
\vspace{-0.2cm}



Many complex systems contain discrete components that interact in continuous time, and maintain interactions that are stochastic, dynamic, and governed by natural laws. For example, molecular systems biology studies molecules (e.g., gene products, proteins) in a living cell that interact according to biochemical laws. An important aspect of studying these systems is predicting the equilibrium behavior of the system upon an intervention, and selecting high-value interventions. For example, we may want to predict the effect of a drug intervention on a new equilibrium of gene expression \cite{alon2006introduction,tyson2003sniffers}. The intervention may have a high value if reduces the expression of a specific gene, while minimizing changes to the other genes.

Recent work in the reinforcement learning community has highlighted the utility of {\it counterfactual policy evaluation} for evaluating and comparing interventions. Counterfactual policy evaluation uses data from past experimental interventions to ask whether a higher value could have been achieved under an alternative intervention \cite{bottou2013counterfactual, joachims2016counterfactual, 2018arXiv181106272B, oberst2019counterfactual}. Counterfactual inference answers this question by predicting the outcome of the alternative intervention, conditional on the outcome of the intervention for which the data were observed  \cite{bottou2013counterfactual, pearl2011algorithmization}.


Predicting the outcome of an intervention requires us to model the system. In particular, {\it discrete-state continuous-time Markov process models} unambiguously describe the changes of system components across all the system states (i.e., not only at equilibrium) in term of hazard functions \cite{Ecological, wilkinson2006stochastic}. A Markov process model predicts the equilibrium upon an intervention by applying the intervention to the initial conditions, performing multiple direct stochastic simulations to reach post-intervention equilibriums, and averaging over these equilibriums. Markov process modeling is one way of modeling complexity in biological systems, particularly in systems that are intrinsically stochastic \cite{alon2006introduction}. The Markov process models are called stochastic kinetic models in this context. 

Unfortunately, Markov process models do not support counterfactual inference. Moreover, it is often impossible to correctly specify a Markov process model of a complex system such as a biological system, where many aspects of the underlying mechanism are unknown. Direct simulations from an incorrectly specified model may incorrectly predict the outcomes of interventions. 

An alternative class of models are \emph{structural causal models} (SCMs). These probabilistic generative causal models are attractive, in that they enable both interventional and counterfactual inference \cite{pearl2009causal}. Recent work used SCMs to model the transition functions in simple Markov decision process models and apply counterfactual policy evaluation to the decisions (i.e. interventions) at each time step \cite{2018arXiv181106272B, oberst2019counterfactual}. Unfortunately, these approaches require outcome data at each time point. This limits their use in situations where we are only interested in the outcome at equilibrium, and only collect data once the equilibrium is reached. 

Defining SCM models at equilibrium directly is non-trivial, because multiple SCMs may be consistent with the equilibrium distribution of the system components upon an intervention, but provide contradictory answers to the same counterfactual query \cite{pearl2009causal, peters2017elements}. Recent work \cite{bongers2018random, mooij2013ordinary, blombeyond} connected a broader class of dynamic models and SCMs, and established the conditions under which interventions in dynamic simulations correspond to SCM's  predictions of equilibrium upon the interventions. However, researchers lack practical examples that leverage this connection, and combine the benefits of these two approaches for counterfactual inference.


This manuscript builds on these prior results, and contributes a general and practical framework for casting the equilibrium distribution Markov process model as an SCM model of equilibrium behavior. The SCMs are defined in terms of the structure and the hazard rates parameters of the Markov process model, and counterfactual inference flows from these settings. The proposed approach alleviates the identifiability drawback of the SCMs, in that their counterfactual inference is consistent with the counterfactual trajectories simulated from the Markov process model. We showcase the benefits of this approach in two studies of cell signal transduction with nonlinear dynamics. The first is a canonical model of the MAPK signaling pathway \cite{KIM2010396}. The second is a larger model that connects the MAPK pathway to stimulus from growth factors \cite{BIANCONI2012142}. We illustrate that, when the underlying Markov process model is misspecified, counterfactual inference anchors intervention predictions to past observed data, and makes selection of interventions more robust to model misspecification.

\vspace{-3mm}
\section{Background}
\vspace{-2mm}
\noindent {\bf Discrete-state continuous-time Markov process models} Discrete-state continuous-time Markov process models describe the temporal interactions between the system components in terms of abstract or physical processes, called {\it rate laws}, with real-valued parameters {\it rates} \cite{Ecological}.  The rate laws determine {\it hazard functions}, which provide instantaneous probabilities of state transitions.

A {\it place invariant} is a set of system components with an invariant sum. A {\it minimal place invariant} can not be further reduced to smaller place invariants \cite{dubins1966invariant}. Define random variables $\mathbf{X}(t) = \{X_i(t): i \in 1 ... J\}$ representing the states of $J$ minimal place invariant components in a Markov process model.  We use capital letters to refer to random variables, lower case letters to refer to instances of random variables, normal font for a single variable, and boldface for a tuple of variables. Denote $P^{\mathbb{M}}(t)$ the probability distribution of $\mathbf{X}(t)$, and $P^{\mathbb{M}}_{X_i}(t)$ the marginal probability of $X_i(t)$. A Markov process model $\mathbb{M}$ is defined by \emph{master equations}, i. e. a coupled set of ordinary differential equations that describe the rate of change of the probabilities of the states $\mathbf{X}(t)$ over time \cite{wilkinson2009stochastic}:
\vspace{-2mm}
\begin{eqnarray}
\frac{dP^{\mathbb{M}}_{X_i}(t)}{dt} = h_i\left(t, \textbf{v}_i, \mathbf{PA}_{\mathbb{M}, i}(t)\right), \  X_i(0) = (x_0)_i \ \forall i \in J
\end{eqnarray}
The function $h_i$ is the hazard function that determines the probability of a state change between $X_i(t)$ and $X_i(s)$, $s>t$. Here $\mathbf{v}_i$ is a set of parameters of the rate laws, and $\mathbf{x}_0$ is an initial condition. $\mathbf{PA}_{\mathbb{M}, i}(t) \subseteq \mathbf{X}(t) \setminus X_i(t)$ is the set of \emph{parents} of variable $X_i(t)$, i.e. variables that regulate $X_i(t)$. Here we consider only Markov process models that converge to unique equilibrium stationary distributions.  If equilibrium exists, then $\lim_{t\to \infty}\frac{dP^{\mathbb{M}}_{X_i}(t)}{dt}=0 $. We denote $X_i^*$ the random variable to which $X_i(t)$ converges in distribution $X_i^* \overset{d}{:=}  \lim_{t\to \infty} X_i$.
We denote $P^{\mathbb{M}}$ the equilibrium distribution of $\mathbf{X}^*$, and $P^{\mathbb{M}}_{X_i^*}$ the marginal probability of $X_i^*$.

\noindent {\bf Equilibrium distribution of a Markov process model as a generative model}
In the equilibrium distribution the place invariants in a Markov process model factorizes into a set of conditional probability distributions, with a causal ordering based on the solutions to the master equations (see Supplementary materials for details). Based on this, the equilibrium distribution can be cast as a causal generative model $\mathbb{G}$ that consists of \cite{pearl2009causal, peters2017elements}:
\vspace{-3mm}
\begin{enumerate}
\item Random variables $\mathbf{X}=\{X_i; i \in 1...J\}$: the states of the system 
\item A directed acyclic graph $\mathbb{D}$ with nodes $\{i \in J\}$ that impose an ordering on $\mathbf{X}$.
\item A set of probabilistic generative functions for each variable $X_i$, $\mathbf{p}=\{p_i, i \in J\}$ such that\\ 
$X_i \sim p_i(\mathbf{PA}_{\mathbb{D}, i}, N_i), \forall i \in J$ 
where $\mathbf{PA}_{\mathbb{D}, i} \subseteq \mathbf{X} \setminus X_i$ are the parents of $X_i$ in $\mathbb{D}$. 
\end{enumerate}
\vspace{-5mm}
$\mathbb{G}$ is a generative model that \emph{entails} an observational distribution $P^{\mathbb{G}}$. This means that a procedure that first samples from each $\mathbf{p_i}$ along the ordering in $\mathbb{D}$ generates samples from $P^{\mathbb{G}}$. This is viewed as the generating process for the observed $\mathbf{X}$. A primary contribution of this work is a method for transforming $\mathbb{G}$ into and structural causal model.

\noindent {\bf Structural causal models (SCMs)} A structural causal model $\mathbb{C}$ of the same system has the same causal directed graph $\mathbb{D}$, ordering the same random variables $\mathbf{X}$. The model consists of \cite{pearl2009causal, peters2017elements}:
\vspace{-3mm}
\begin{enumerate}
\item A distribution $P_{\mathbf{N}}^{\mathbb{C}}$ on independent \emph{noise} random variables $\mathbf{N} = \{N_i; i \in J \}$ \vspace{-3mm}
\item A set of functions $\mathbf{f}=\{f_i, i \in J\}$ called \emph{structural assignments}, such that\\
$X_i = f_i(\mathbf{PA}_{\mathbb{C}, i}, N_i), \forall i \in J$
where $\mathbf{PA}_{\mathbb{C}, i} \subseteq \mathbf{X} \setminus X_i$ are the parents of $X_i$ in $\mathbb{D}$.
\end{enumerate}
$\mathbb{C}$ is a generative model that entails $P^{\mathbb{G}}$, the same observational distribution as $\mathbb{G}$.  For consistency, we refer to this distribution as $P^{\mathbb{C}}$ when discussed in the context of $\mathbb{C}$.   This means that a procedure that first samples noise values from $P_{\mathbf{N}}^{\mathbb{C}}$, and then sets the values of $\mathbf{X}$ deterministically with $\mathbf{f}$, generates samples from $P^{\mathbb{C}}$. This is viewed as the generating process for the observed $\mathbf{X}$.

\noindent {\bf Interventions in Markov process models and in SCMs} An SCM $\mathbb{C}$ uses {\it ideal interventions}, which replace a random variable with a fixed point value. These are represented with Pearl's ``do'' notation $\text{do}(X_i = x)$ \cite{doi:10.1086/525638, pearl2019interpretation}, denoted. The intervention that sets $X_i$ to $x$ replaces the structural assignment $X_i = f_i(\mathbf{PA}_{\mathbb{C}, i}, N_i)$ with $X_i = x$.  The \emph{intervention distribution} $P^{\mathbb{C}; \text{do}(X_i = x)}$ is entailed by $\mathbb{C}$ under the intervention and is generally different from the equilibrium distribution $P^{\mathbb{M}}$ of $\mathbf{X}^*$.

In the context of a Markov process model, a typical intervention definition is that an intervention increases a reaction rate (catalyzation) or decreases a reaction rate (inhibition). We define a type of {\it soft intervention} \cite{doi:10.1086/525638} for Markov process models that make this rate manipulation comparable to the SCM's ideal intervention.  We define a fixed post-equilibrium expected value for a variable that we want to achieve, then find a change to the variables rate parameter values that achieve that outcome.  For example, an intervention that sets the equilibrium value of $X_i$ to $x$ does so by finding manipulating $X_i$'s rate parameters to achieve this result. Borrowing the ``do'' notation, denote this as $\text{do}(X_i^* = x)$. Let the equilibrium distribution under intervention be $P^{\mathbb{M}; \text{do}(X_i^* = x)}$. We compare intervention queries on $P^{\mathbb{M}; \text{do}(X_i^* = x)}$ to $P^{\mathbb{C}; \text{do}(X_i = x)}$. For both Markov process models and SCMs, the intervention queries are answered by sampling from these distributions.  See Supplementary materials for contrasts to related intervention modeling approaches.

\noindent {\bf Counterfactual inference in SCMs} Counterfactual inference is the process of observing a random outcome, making inference about the unseen causes of the outcome, and then inferring the outcome that would have been observed under an intervention \cite{peters2017elements, roese1997counterfactual}. For example, an SCM $\mathbb{C}$ helps answer the query ``\emph{Having observed $X_i = x$, what would have happened under the intervention $\text{do}(X_i = \neg x)$?}". SCMs support the following algorithm for counterfactual inference  \cite{Balke:1994:CPC:2074394.2074401}: (1) having observed $X = x$, infer the noise distribution conditional on the observation $P_{\mathbf{N}}^{\mathbb{C}; X=x}$, (2) replace $P_{\mathbf{N}}^{\mathbb{C}}$ with $P_{\mathbf{N}}^{\mathbb{C}; X=x}$ in $\mathbb{C}$, (3) apply the intervention $\text{do}(X = \neg x)$, and (4) sample from the resulting mutated model.  The intuition is that in (2) we infer the latent initial conditions (values of $N$) that could have lead to the outcome $X = x$, this information is encoded in $P_{\mathbf{N}}^{\mathbb{C}; X=x}$, the posterior of $N$ given $X = x$.  We then pass that encoded information to the counterfactual world where $X$ is set to $\neg x$ and play out scenarios in that world by sampling from $P_{\mathbf{N}}^{\mathbb{C}; X=x}$ and deriving downstream variables given those noise values. Thus the algorithm mutates $\mathbb{C}$ into an SCM entailing the \emph{counterfactual distribution} $P^{\mathbb{C}; X = x, \text{do} (X = \neg x)}$. 

\section{Methods}
\subsection{Motivating example}

This manuscript contributes a practical framework for casting Markov process models of a system observed at equilibrium as an SCM, for the purposes of conducting counterfactual inference. As a motivating example, we consider a system of three biomolecules (i.e., components) $\mbox{X}_1$, $\mbox{X}_2$ and $\mbox{Y}$. Each component takes two states: active (``on") and inactive (``off"). Component $\mbox{X}_1$ in the ``on" state activates $\mbox{Y}$; component $\mbox{X}_2$ in the ``on" state deactivates $\mbox{Y}$, as shown in the causal diagram \cite{alon2006introduction} below:
\begin{eqnarray}
\mbox{X}_1^{\text{on}} + \mbox{Y}^{\text{off}}   &\overset{v_1}{\rightarrow} \ \mbox{X}_1^{\text{on}} + \mbox{Y}^{\text{on}} \label{e1}\  \ \mbox{and}\ \
\mbox{X}_2^{\text{on}} + \mbox{Y}^{\text{on}} &\overset{v_2}{\rightarrow} \ \mbox{X}_2^{\text{on}} + \mbox{Y}^{\text{off}}
\end{eqnarray}
Let $X_1(t)$, $X_2(t)$, and $Y(t)$ be the total number of active-state particles of $\mbox{X}_1$, $\mbox{X}_2$, and $\mbox{Y}$  at time $t$. Assume that each component has $T=100$ particles in total, such that $T - Y(t)$ is the number of inactive particles of $\mbox{Y}$ at time $t$, and that each component is initialized with 100 off-state particles. 

To ensure that the equilibrium distribution of the Markov process model $\mathbb{M}$ has a closed-form solution, we limit this work to $\mathbb{M}$ with zero or first-order hazard functions (i.e. hazard functions for which outputs are either constant or directly proportional to a product of the inputs) \cite{jahnke2007solving, wilkinson2009stochastic}. In this example, the hazard functions assume {\it mass action kinetics} \cite{horn1972general}, a common assumption in biochemical modeling.  Let $h_1(Y(t))$ and $h_2(Y(t))$ denote stochastic rate laws for the activation and deactivation of $\mbox{Y}$, expressing the probabilities that the reactions occur in the instant $(t, t+dt]$. Then, according to a first-order stochastic kinetic assumption of chemical reactions \cite{wilkinson2006stochastic}, $h_1$ and $h_2$ are
\begin{eqnarray}
h_1(Y(t)) = v_1 X_1(t) (T- Y(t)) \ \ \mbox{and} \ \ h_2(Y(t)) = v_2 X_2(t)Y(t) \label{toy:hazard}
\end{eqnarray}
The hazard functions are parameterized by $\mathbf{v}=\{v_1,v_2\}$ regulating $\mbox{X}_1$ and $\mbox{X}_2$, and by the initial states. 

The Kolmogorov forward equations determine the change in $P^{\mathbb{M}}_{Y(t)}$ as the system evolves in time:
\vspace{-.1in}
\begin{eqnarray}
& \frac{dP^{\mathbb{M}}_{Y(t)}}{dt} = \left (h_1(Y(t) - 1)P^{\mathbb{M}}_{Y(t)-1} - h_1(Y(t))P^{\mathbb{M}}_{Y(t)} \right ) +  \left (h_2(Y(t) + 1)P^{\mathbb{M}}_{Y(t)+1} - h_2(Y(t))P^{\mathbb{M}}_{Y(t)} \right ) &   \label{dyn_prob}
\end{eqnarray}
We pose a counterfactual query ``\emph{Having observed $X_1 = 34, X_2 = 45, Y = 56$, what would $Y$ have been if $X_1$ was set to 50}"?

\subsection{Converting a Markov process model into an SCM \label{sec:SCM}}

Algorithm \ref{SCM} summarizes the proposed steps of converting the Markov process model into an SCM. The steps are a series of mathematical derivations  (as opposed to a pseudocode for a computational implementation). Below we illustrate these steps for the component $\mbox{Y}$ in the motivating example. Additional mathematical details are available in Supplementary materials.

\vspace{-3mm}

\noindent
\begin{minipage}[t!]{.46\textwidth}
\raggedright

\begin{algorithm}[H]
\begin{small}
\caption{ \small \textit{Convert Markov process into SCM}
\newline Inputs: $~~~$ Markov process model $\mathbb{M}$
\newline Output: $~~~$ Structural causal model $\mathbb{C}$
}

\label{SCM}
\begin{algorithmic}[1]
\Procedure{GetSCM}{$\mathbb{M}$}
\LineComment{{\footnotesize Solve master equation}}
\State $P^{\mathbb{M}}(t) :=\int_t \frac{dP^{\mathbb{M}}(t)}{dt}) dt$ \label{master_equation}
\LineComment{{\footnotesize Find the equilibrium distribution}}
\State {$P^{\mathbb{M}}  =\lim_{t\to \infty} P^{\mathbb{M}}(t) $} \label{find_steady_state}
\LineComment{{\footnotesize Use $ P^{\mathbb{M}} $ to define generative model $\mathbb{G}$}}
\State {$\mathbb{G}:= \{x \sim P^{\mathbb{M}} \}$} \label{build_pgm}
\LineComment{{\footnotesize Convert the generative model to an SCM }}
\LineComment{{\footnotesize that entails $P^{\mathbb{M}}$}}
\State {
$\mathbb{C}:= \left\{\begin{matrix}
\mathbf{N} \sim P_{\mathbf{N}}^{\mathbb{C}} & \\
\mathbf{X} = \mathbf{f}(\mathbf{X}, \mathbf{N}) &
\end{matrix}\right. : P^{\mathbb{C}} \approx P^{\mathbb{M}}$
} \label{build_scm}
\State \textbf{return} {$\mathbb{C}$}
\EndProcedure
\end{algorithmic}
\end{small}
\end{algorithm}


\end{minipage}
\begin{minipage}[t!]{.53\textwidth}
\raggedleft

\begin{algorithm}[H]
\begin{small}
\caption{ \small \textit{Counterfactual inference on SCM}
\newline Inputs: Prior distribution on exogenous noise NPrior
\newline $~~~~~~~~~~~~$ Structural causal model  $\mathbb{C}$
\newline $~~~~~~~~~~~~$ Observed endogenous variables $X=x$
\newline $~~~~~~~~~~~~$ Counterfactual interventions $X=\neg x$
\newline $~~~~~~~~~~~~$ Desired sample size $ssize$
\newline Output: $ssize$ samples from $P^{\mathbb{C}; X=x, do(X:=\neg x)}$ }

\label{CF}
\begin{algorithmic}[1]
\Procedure{CFQuery}{$\mathbb{C}$, NPrior, $x$, $ \neg x$, $ssize$}
\LineComment{{\footnotesize Create  ``observation" and ``intervention"  models}}
\State {$\textbf{obsModel}$ $\gets$ Condition($\mathbb{C}$, $X=x$)} \label{obsModel}
\State {$\textbf{intModel}$ $\gets$ Do($\mathbb{C}$, $X = \neg x$) } \label{metaprog}
\LineComment{{\footnotesize Infer noise distribution with observation model}}
\State  {NPosterior $\gets$ Infer($\textbf{obsModel}$, NPrior)} \label{update_noise}
\LineComment{{\footnotesize Simulate from intervention model w/ updated noise}}
\State samples = array($ssize$)
\For{i in (0:$ssize$)}
\State{ samples[i] $\gets$ $\textbf{intModel}$(NPosterior)}\label{simulate_cf}
\EndFor
\State {\textbf{return} samples}
\EndProcedure
\end{algorithmic}
\end{small}
\end{algorithm}

\end{minipage}

\noindent {\bf Solve the master equation} (Algo. \ref{SCM} line \ref{master_equation}).
We can arrive at the solution for $P^{\mathbb{M}}_{Y(t)}$  in \eqref{dyn_prob} indirectly by solving the ordinary differential equation on the expectation of $Y(t)$ over $P^{\mathbb{M}}_{Y(t)}$:
\vspace{-3mm}
\begin{eqnarray}
\frac{d}{dt}E(Y(t)) = v_1X_1(t)T - \left(v_1X_1(t) + v_2X_2(t)\right)  E(Y(t)) \label{dE}
\end{eqnarray}
This has an analytical solution, where:
\vspace{-2mm}
\begin{eqnarray}
\frac{E(Y(t))}{T} =  e^{-t(v_1X_{1}(t) + v_2X_{2}(t))} + \frac{v_1 X_1(t)}{v_1 X_1(t) + v_2 X_2(t)}\label{dyn_theta}
\end{eqnarray}
Finally, $Y(t)$ is a count of binary state variables with the same probability of being activated at a given instant. Then  $P^{\mathbb{M}}_{Y(t)}$ must be Binomial distribution with $T$ trials, and trial probability $\frac{E(Y(t))}{T}$.

\noindent {\bf Find the equilibrium distribution} (Algo. \ref{SCM} line \ref{find_steady_state}).
Taking the limit in time of \eqref{dyn_theta}: \begin{align}
\frac{E(Y)}{T} = \underset{t \to \infty}{lim} \frac{E(Y(t))}{T} = \frac{v_1 X_1(t)}{v_1 X_1(t) + v_2 X_2(t)}
\label{get_equilibrium}\end{align} Thus at equilibrium $Y$ follows the Binomial probability distribution with parameter $\frac{v_1 X_1(t)}{v_1 X_1(t) + v_2 X_2(t)}$. 

\medskip \noindent {\bf Use $ P^{\mathbb{M}} $ to define generative model $\mathbb{G}$ } (Algo. \ref{SCM} line \ref{build_pgm}). Let $\theta_{X_1}$ and $\theta_{X_1}$ be the probability parameters for the equilibrium Binomial distributions $P^{\mathbb{M}}_{X_1}$ and $P^{\mathbb{M}}_{X_2}$. Let
$\theta_{Y}(X_1, X_2) = \frac{E(Y)}{T}$ be the probability parameter for the equilibrium Binomial distribution $P^{\mathbb{M}}_{Y}$.
Define a generative model $\mathbb{G}$:
\begin{eqnarray}
\mathbb{G}:= \left\{X_1 \sim \text{Binom}(T, \theta_{X_1}); \ X_2 \sim \text{Binom}(T, \theta_{X_2});\ 
Y \sim \text{Binom}(T, \theta_{Y}(X_1, X_2)) \right\} \label{toy_gen_model}
\end{eqnarray}

\noindent {\bf Convert the generative model to an SCM that entails $P^{\mathbb{M}}$ } (Algo. \ref{SCM} line \ref{build_scm}).
We rely on a method of {\it monotonic conversion}, which restricts the class of possible SCMs to those with a common set of identifiable counterfactual quantities (such as the {\it probability of necessity}, i.e. the probability that $\mbox{Y}$ would not have been activated without $\mbox{X}_1$) \cite{pearl2009causal}. For each structural assignment $X_i = f_i(\mathbf{PA}_{\mathbb{C}, i}, N_i), \forall i \in J$ the method enforces the property  $E[X_i \mid do(\mathbf{PA}_{\mathbb{C}, i} = y)] \ge E[X_i \mid do(\mathbf{PA}_{\mathbb{C}, i}=y^{\prime})] \Rightarrow f_i(y, n_i) \ge f_i(y^{\prime},n_i) \forall n_i$.  

For this example we selected a monotonic conversion by means of the inverse CDF transform. Denote $F^{-1}(u, n, p)$ the inverse CDF of the Binomial distribution, where $0< u <1$, and $n$ (number of trials) and $p$ (success probability) are the parameters of the Binomial distribution.
Then the SCM $\mathbb{C}$ that entails $P^{\mathbb{M}}$ is defined as
\vspace{-.15in}
\begin{eqnarray}
& N_{X_1}, N_{X_2}, N_{Y} \overset{\text{ind}}{\sim} \text{Uniform}(0, 1); &\\ 
& \mathbb{C} := \left\{
X_1 = F^{-1}(N_{X_1}, T, \theta_{X_1});\ 
X_2 =  F^{-1}(N_{X_2}, T, \theta_{X_2});  \ 
Y = F^{-1}(N_{Y}, T,\theta_{Y}(X_1, X_2)) \right\}  \nonumber &
\label{toy_scm_model}
\end{eqnarray}
For larger models such as in Case studies 1 and 2 thereafter, it may be desirable to work with alternative transforms that are more amenable to gradient-based inference such as stochastic variational inference.


\subsection{Counterfactual inference and evaluation} 

Algorithm \ref{CF} details the counterfactual inference on $\mathbb{C}$. Algorithms 3 and 4 in Supplementary materials detail the evaluation. The evaluation stems from the insight that noise at the equilibrium captures the stochasticity in the Markov process trajectories. Therefore, we repeatedly simulate pairs of the trajectories with and without the counterfactual intervention, with a same random seed in a pair, such that each pair has an identical stochastic component. We then compare the differences in the values of these pairs at equilibrium to the differences between the original and the intervened-upon values projected by the SCM. These differences estimate the respective {\it causal effects}. The algorithms differ in choosing a deterministic or a stochastic approach for the estimation of causal effects. To ensure scalability to large models and the ability to do inference over a broad set of structural assignments, we implemented the algorithms in PyTorch and the probabilistic programming language {\it Pyro} \cite{bingham2018pyro}. The code and the runtime data are in Supplementary materials.


\section{Case studies}

\subsection{Case Study 1: The MAPK signaling pathway \label{sec:mapk_def}}

{\bf The system} The mitogen-activated protein kinase (MAPK) pathway is important in many biological processes, such as determination of cell fate. It is a cascade of three proteins, a MAPK, a MAPK kinase (MAP2K), and a MAPK kinase kinase (MAP3K), represented with a causal diagram \cite{huang1996,qiao2007bistability}
\begin{eqnarray}
& \text{E1} \rightarrow \text{MAP3K} \rightarrow \text{MAP2K} \rightarrow \text{MAPK} & \label{eq:MAPKdef}
\end{eqnarray}
Here E1 is an input signal to the pathway. The cascade relays the signal from one protein to the next by changing the count of proteins in an active state.

{\bf The biochemical reactions} A protein molecule is in an active state if it has one or more attached phosphoryl groups. Each arrow in \eqref{eq:MAPKdef} combines the reactions of  phosphorylation (i.e., activation) and dephosphorylation (i.e., desactivation). For example, $\text{E1} \rightarrow \text{MAP3K}$ combines two reactions \vspace{-2mm}
\begin{eqnarray}
\text{E1} + \text{MAP3K} \overset{v^{act}_{K3}}{\rightarrow} \text{E1} + \text{P-MAP3K} \ \mbox{and}\
\text{P-MAP3K} \overset{v^{inh}_{K3}}{\rightarrow} \text{MAP3K} \label{eq:MAPKrates}
\vspace{-2mm}
\end{eqnarray}

\vspace{-.1in}
In the first reaction in~\eqref{eq:MAPKrates}, a particle of the input signal \text{E1} binds (i.e., activates) a molecule of MAP3K to produce MAP3K with an attached phosphoryl. The rate parameter associated with this reaction is $v^{act}$. In the second reaction, phosphorylated MAP3K loses its phosphoryl (i.e., deactivates), with the rate $v^{inh}$.  The remaining arrows in \eqref{eq:MAPKdef} aggregate similar reactions and rate pairs.

{\bf The mechanistic model} Let $\text{K3}(t)$, $\text{K2}(t)$ and $\text{K}(t)$ denote the counts of phosphorylated \text{MAP3K}, \text{MAP2K}, and \text{MAPK} at time $t$. Let $T_{\text{K3}}$, $T_{\text{K2}}$, and $T_{\text{K}}$ represent the total amount of each of the three proteins, and $E1$ the total amount of input, which we assume are constant in time. We model the system as a continuous-time discrete-state Markov process $\mathbb{M}$ with hazard rates functions in Table \ref{dyn_model}. 

{\bf The data} We simulated the counts of protein particles using the Markov process model with rate parameters $v^{act}_{K3}$, $v^{act}_{K2}$, $v^{act}_{K}$ and deactivation rate parameters $v^{inh}_{K3}$, $v^{inh}_{K2}$, $v^{inh}_{K}$. We conducted three simulation experiments with three sets of rates, all consistent with a low concentration in a cell-sized volume (see Supplementary materials). The initial conditions assumed 1 particle of \text{E1}, 100 particles of the unphosphorylated form of each protein, and 0 particles of the phosphorylated form.

{\bf The counterfactual of interest} Let $\text{K3}$, $\text{K2}$ and $\text{K}$ denote the observed counts of phosphorylated \text{MAP3K}, \text{MAP2K}, and \text{MAPK} at 100 seconds, the time corresponding to an equilibrium for all the rates.  Let K3$^{\prime}$ be the count of phosphorylated MAP3K generated by a 3 times smaller $v^{act}_{K3}$. Thus $\mathbf{v}^{\prime} = [v^{act}_{K3}/3, v^{inh}_{K3},v^{act}_{K2}, v^{inh}_{K2}, v^{act}_{K}, v^{inh}_{K}]$.   We pose the counterfactual question: \emph{``Having observed the equilibrium particle counts $\text{K3}$, $\text{K2}$ and $\text{K}$, what would have been the count of $\text{K}$ if we had $\text{K3}^{\prime}$?''}. 

{\bf The evaluation} We derive the SCM $\mathbb{C}$ of the Markov process model and evaluate the counterfactual distribution $P_{\text{K3}}^{\mathbb{C}; \text{K3}=\mathbf{x}, \text{K2}=\mathbf{y}, \text{K}=\mathbf{z}, \text{do}(\text{K3} = x^{\prime})}$ where $x^{\prime}$ is the expected equilibrium value associated with $v^{\prime}$. We evaluate this counterfactual statement as described in Algorithms 3 and 4 (with 500 seeds). If the counterfactuals from the converted SCMs are consistent with the Markov process models, their histograms from Algorithms 3 and 4 should overlap.

{\bf The evaluation under model misspecification} We consider the Markov process model $\mathbb{M}$ with the first set of rates (see Supplementary materials). Let $[\mathbf{x},\mathbf{y},\mathbf{z}]$ be sampled from $\mathbb{M}$.
Next, instead of the correct model we consider a misspecified model $\mathbb{M}^{\prime}$, where $v^{act}_{K2}$ is perturbed with noise sampled from Uniform$(0.1, 0.5)$.
We denote as $\mathbb{C}^{\prime}$  the SCM corresponding to $\mathbb{M}^{\prime}$, and evaluate the counterfactual distribution $P_{\text{K3}}^{\mathbb{C}^{\prime}; \text{K3}=\mathbf{x}, \text{K2}=\mathbf{y}, \text{K}=\mathbf{z}, \text{do}(\text{K3} = x^{\prime})}$. We expect that, since the counterfactual distribution from $\mathbb{C}^{\prime}$ incorporates the data from the correct model, it should be closer to the true causal effect simulated from $\mathbb{M}$ than the direct simulation from the misspecified $\mathbb{M}^{\prime}$. We repeat this experiment 50 times.
\begin{table}[]
\addtolength{\parskip}{-3mm}
\centering
\scalebox{0.75}{
\begin{tabular}{l|lll}
                    & MAP3K                                                     & MAP2K                                                     & MAPK                                                     \\ \hline
activation hazard   & $v^{act}_{K3}\text{E1}(T_{\text{K3}}-\text{K3}(t))$ & $v^{act}_{K2}\text{K3}(T_{\text{K2}}-\text{K2}(t))$ & $v^{act}_{K}\text{K2}(T_{\text{K}}-\text{K}(t))$ \\
deactivation hazard & $v^{inh}_{K3}\text{K3}(t)$                       & $v^{inh}_{K2}\text{K2}(t)$                       & $v^{inh}_{K}\text{K}(t)$
\end{tabular}
}
\caption{\small The hazard functions in Case study 1 (MAPK), specified according to mass action enzyme kinetics. \label{dyn_model}}

\vspace{-5mm}
\end{table}

\subsection{Case Study 2: The IGF signaling system}

{\bf The system} The growth factor signaling system is involved in growth and development of tissues. When external stimuli activate the epidermal growth factor (EGF) or the insulin-like growth factor (IGF), this triggers a cascade \cite{BIANCONI2012142} in \figref{fig:igf_ce}(a). The Raf-Mek-Erk pathway is equivalent to \eqref{eq:MAPKdef},  renamed to follow the convention adopted by the biological literature in this context.

{\bf The biochemical reactions} All the edges in \figref{fig:igf_ce}(a) represent enzyme reactions E + S $\overset{v}{\rightarrow}$ E + P, where the change of substrate S to product P is catalyzed by enzyme E. As in Case study 1, the pointed edges combine activation and deactivation. The flat-headed edges only represent deactivation. The mechanistic model is built similarly to Case study 1.

{\bf The data} We simulated the counts of protein particles using the Markov process model with rates in Supplemental Tables 2 and 3. The other settings are as in Case study 1. The initial condition assumed 37 particles of EGFR, 5 particles of IGFR, 100 particles of the unphosphorylated form of other proteins, and 0 particles of the phosphorylated form.

{\bf The counterfactual of interest} Let $R^{\prime}$ be the number of phosphorylated particles of $\text{Ras}$ at equilibrium, achieved with $v_{\text{Ras-SOS}}^{\prime \text{act}} = v_{\text{Ras-SOS}}^{\text{act}}/6$. We pose the counterfactual: \emph{Having observed the number of phosphorylated particles of each protein before the intervention, what would be the number of particles of Erk if the intervention had fixed $\text{Ras}=\text{R}^{\prime}$?} Unlike the MAPK pathway, where the  intervention on \text{MAP3K} affects the counterfactual target \text{MAPK} through a direct path, this system has two paths from Ras to Erk. One path goes directly through Raf, and the other through a mediating path PI3K $\rightarrow$ AKT. This challenges the algorithm to address multiple paths of influence.

{\bf The evaluation} We consider the rates $v_{\text{Ras-SOS}}^{\text{act}}/6$, the counterfactual distribution $P_{\text{Erk}}^{\mathbb{C}; X_i = x_i, \text{do}(\text{Ras} = R^{\prime})}$, and the Algorithms 3 and 4 (with 300 seeds).

{\bf The evaluation under model misspecification} We consider Markov process model $\mathbb{M}$ with the same rates and initial conditions as above. Let $x_i$ be sampled from $\mathbb{M}$.
We then introduce a misspecified model $\mathbb{M}^{\prime}$, where $v^{act}_{\text{AKT-PI3K}}$ is perturbed with noise sampled from Uniform$(0.01, 0.1)$.
We denote as $\mathbb{C}^{\prime}$  the SCM corresponding to $\mathbb{M}^{\prime}$, and evaluate the counterfactual distribution $P_{\text{Erk}}^{\mathbb{C^{\prime}}; X_i = x_i, \text{do}(\text{Ras} = R^{\prime})}$.
The resulting counterfactual distribution from $\mathbb{C}^{\prime}$ should be closer to the true causal effect simulated from $\mathbb{M}$ than the direct simulation from the misspecified $\mathbb{M}^{\prime}$. We repeat this experiment 50 times.

\vspace{-4mm}
\begin{figure}[htbp!]
\begin{center}
\begin{tabular}{ccc}
\includegraphics[width=.22\textwidth]{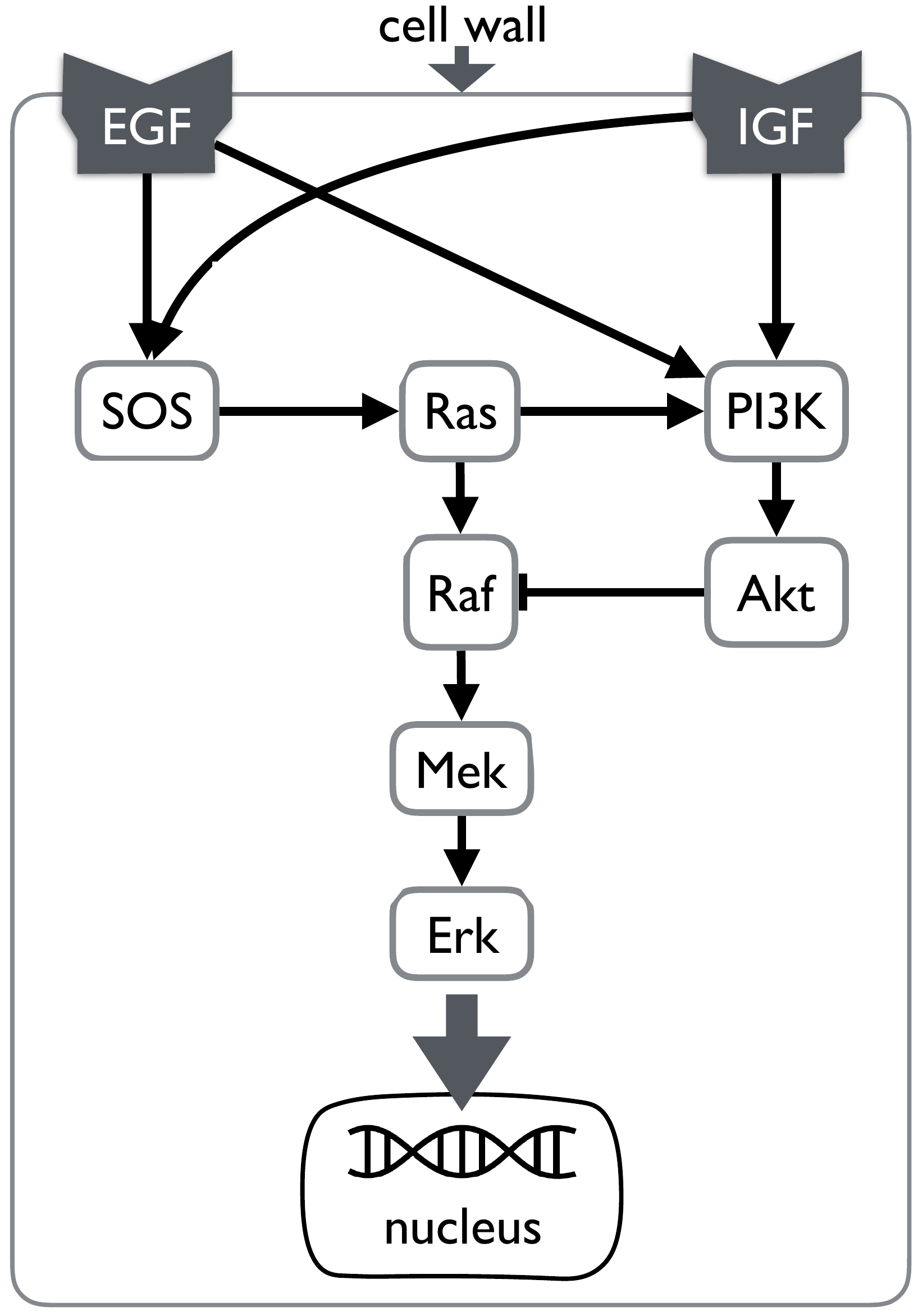} &
\includegraphics[width=0.4\textwidth]{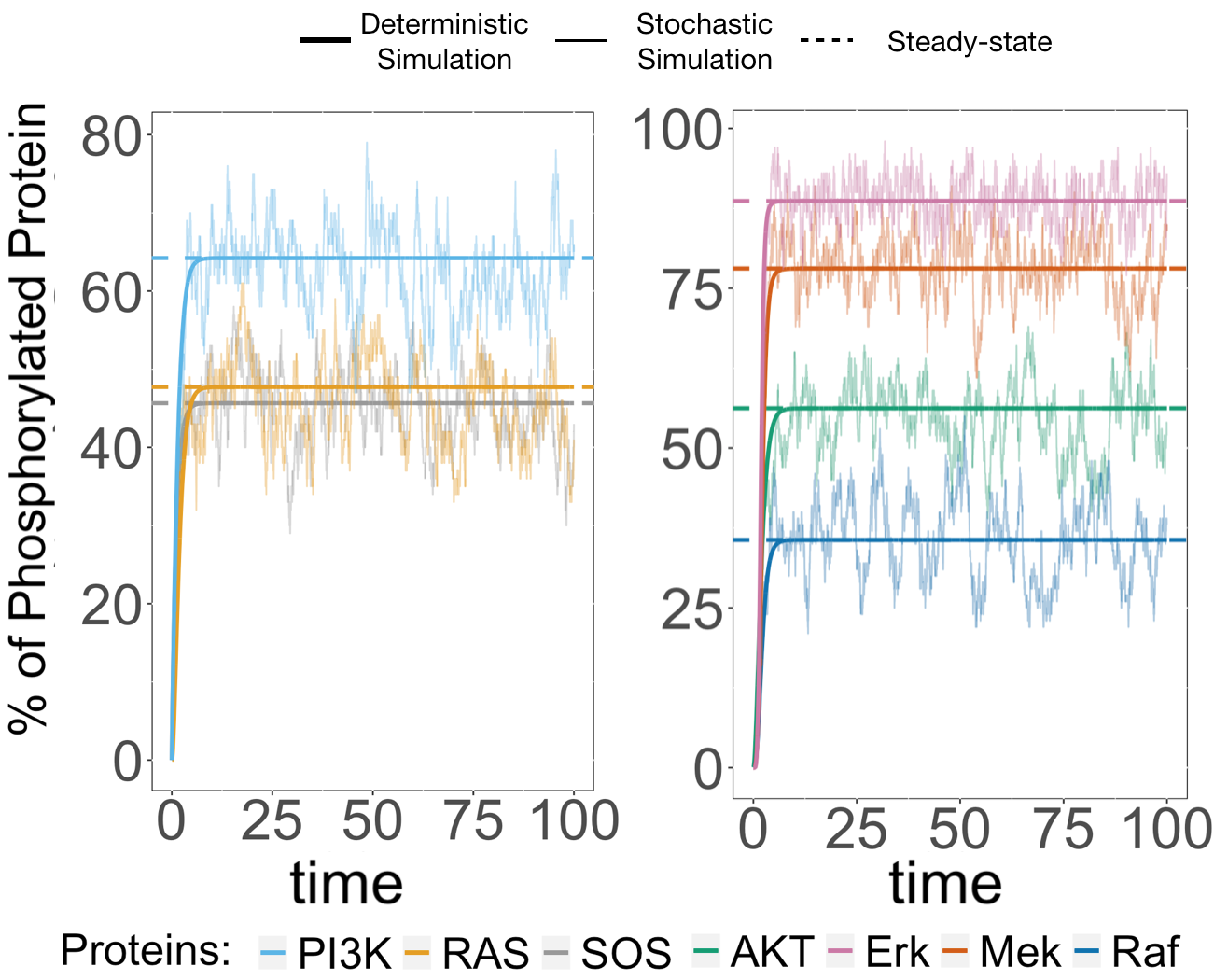}\label{fig:IGFSteadyState} &
\includegraphics[width=0.28\textwidth]{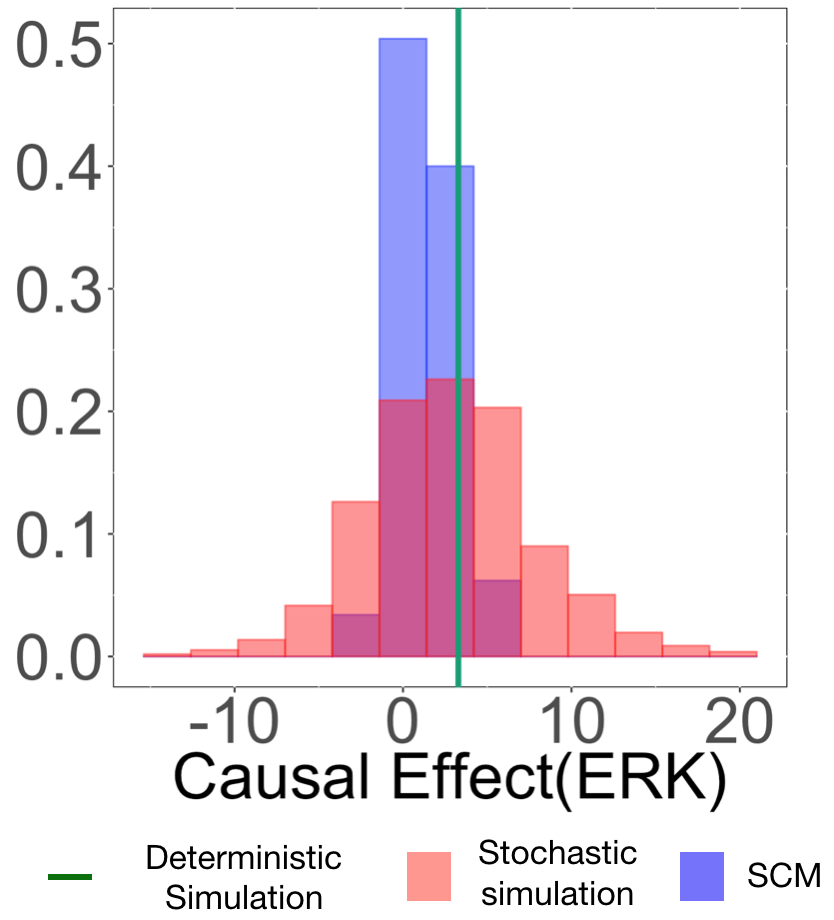}\label{fig:igf_ce}\\
(a) & (b) & (c)
\end{tabular}
\end{center}
\vspace{-.15in}
    \caption{\small \textbf{Case study 2 (IGF).} (a) IGF signaling.
The top nodes are receptors for the epidermal growth factor (EGF) and the insulin growth factor (IGF). Downstream of the receptors are several canonical signaling pathways (including Raf-Mek-Erk, a renamed equivalent of \eqref{eq:MAPKdef}). Each reaction has a single rate parameter.  The auto-deactivation reactions are not pictured. (b) Deterministic and stochastic trajectories of the active-state proteins in the system. Horizontal lines are the expected values at equilibrium. (c) Histogram of causal effects, defined as differences between the ``observed" and the ``counterfactual" trajectories of $\text{ERK}$ at equilibrium.}  \label{fig:igf_ce}
\end{figure}


\vspace{-4mm}
\section{Results}

\vspace{-2mm}
\subsection{Case Study 1: The MAPK signaling pathway}

\noindent {\bf Solve stochastic process's master equation} (Algorithm \ref{SCM} line \ref{master_equation}).
As in the motivating example, we indirectly solve $\frac{dP^{\mathbb{M}}(t)}{dt}$ by way of the solving the forward equations for the expectation.  For $\text{K3}(t)$ this is $\frac{dE(\text{K3}(t))}{dt} = v^{act}_{\text{K3}}\text{E1}(T_{\text{K3}}-E({\text{K3}(t)}) - v^{inh}_{\text{K3}}E(\text{K3}(t)))$ (Added Expectation in RHS, please review). We derive similar forward equations for $\text{K2}( t)$ and $\text{K}( t)$.
We solve the ODE above:
 \begin{align}
\frac{E(\text{K3}(t))}{T_{\text{K3}}} &= e^{-t(v^{act}_{\text{K3}}\text{E1} + v^{inh}_{\text{K3}})} + \frac{v^{act}_{\text{K3}}\text{E1}}{v^{act}_{\text{K3}}\text{E1} + v^{inh}_{\text{K3}}}
 \label{3K_solution}
 \end{align}
and obtain the equilibrium by taking the limit $t \rightarrow \infty$. The first term in \eqref{3K_solution} goes to 0:
 \begin{align}
\frac{E(\text{K3})}{T_{\text{K3}}} &= \frac{v^{act}_{\text{K3}}T_{\text{K3}}\text{E1}}{v^{act}_{\text{K3}}\text{E1} + v^{inh}_{\text{K3}}}
 \label{3K_ss}
 \end{align}

\noindent {\bf Find the equilibrium distribution} (Algorithm \ref{SCM} line \ref{find_steady_state})
As in \secref{sec:SCM}), the each active-state MAPK protein has a Binomial marginal distribution.
Let $\theta_{\text{K3}}(\text{E1})$ denote the probability that a MAP3K particle is active at equilibrium given \text{E1}.
After solving the master equation,
\vspace{-2mm}
\begin{align}
\theta_{\text{K3}}(\text{E1}) &= \frac{E(\text{K3})}{T_{\text{K3}}}=\frac{v^{act}_{\text{K3}}\text{E1}}{v^{act}_{\text{K3}}\text{E1} + v^{inh}_{\text{K3}}} \label{3K_equiv}
\end{align}
 Extending this solution to MAP2K and MAPK leads to probabilities
 \vspace{-2mm}
\begin{eqnarray}
\theta_{\text{K3}}(\text{E1}) = \frac{v^{act}_{\text{K3}}\text{E1}}{v^{act}_{\text{K3}}\text{E1} + v^{inh}_{\text{K3}}} ; \ 
\theta_{\text{K2}}(\text{K3}) = \frac{v^{act}_{\text{K2}}\text{K3}}{v^{act}_{\text{K2}}\text{K3} + v^{inh}_{\text{K2}}} ; \ 
\theta_{\text{K}}(\text{K2}) = \frac{v^{act}_{\text{K}}\text{K2}}{v^{act}_{\text{K}}\text{K2} + v^{inh}_{\text{K}}} \label{K_equilibrium}
\end{eqnarray}
and the following equilibrium distributions:
\vspace{-2mm}
\begin{eqnarray}
P_{\text{K3}}^{\mathbb{M}} \equiv \text{Binomial}(T_{\text{K3}}, \theta_{\text{K3}}(\text{E1})); \
P_{\text{K2}}^{\mathbb{M}} \equiv \text{Binomial}(T_{\text{K2}}, \theta_{\text{K2}}(\text{K3})); \
P_{\text{K}}^{\mathbb{M}} \equiv \text{Binomial}(T_{\text{K}}, \theta_{\text{K}}(\text{K2})) \label{binom_k}
\end{eqnarray}

\vspace{-2mm}
\noindent {\bf Use $ P^{\mathbb{M}} $ to define generative model $\mathbb{G}$} (Algorithm \ref{SCM} line \ref{build_pgm}).
From here it is straightforward to create a generative model that entails $P_{\text{K3}}^{\mathbb{M}}$:
\begin{eqnarray}
\mathbb{G}:= \left\{
\text{K3} \sim \text{Binom}(T_{\text{K3}}, \theta_{\text{K3}}(\text{E1})); \ 
\text{K2} \sim \text{Binom}(T_{\text{K2}}, \theta_{\text{K2}}(\text{K3})); \ 
\text{K} \sim \text{Binom}(T_{\text{K}}, \theta_{\text{K}}(\text{K2}))
\right\} \label{gen_model}
\end{eqnarray}

\vspace{-2mm}
\noindent {\bf Convert the generative model to an SCM that entails $P^{\mathbb{M}}$} (Algorithm \ref{SCM} line \ref{build_scm}). Here the challenge is in expressing the stochasticity in $\mathbb{G}$, while defining \text{K3}, \text{K2}, \text{K} as deterministic functions of the noise variables $N_{\text{K}}, N_{\text{K2}}, N_{\text{K3}}$. Instead of using the inverse binomial CDF, we demonstrate the use of a differentiable monotonic conversion, so that we can validate approximate counterfactual inference with stochastic gradient descent.  We achieve this by first applying a Gaussian approximation to the Binomial distribution, and then applying the ``reparameterization trick" used in variational autoencoders \cite{rezende2014stochastic} (combined in helper function $q$ in \eqref{trick}).  
\vspace{-2mm}
\begin{eqnarray}
& N_{\text{K}}, \ N_{\text{K2}}, N_{\text{K3}} \overset{\text{ind}}{\sim} N(0, 1);\  q(\theta, T, N) = N \cdot (T \theta (1 - \theta))^{1/2} + \theta T & \label{trick}\\
& \mathbb{C}:= \left\{
\text{K3} = q(\theta_{\text{K3}}(\text{E1}), T_{\text{K3}}, N_{\text{K3}});\ 
\text{K2} =  q(\theta_{\text{K2}}(\text{K3}), T_{\text{K2}}, N_{\text{K2}});\ 
\text{K} = q(\theta_{\text{K}}(\text{K2}), T_{\text{K}}, N_{\text{K}}) \right\}~~~
 & \label{scm_model}
\end{eqnarray}

The Gaussian approximation facilitates the gradient-based inference in line \ref{update_noise} of Algorithm \ref{CF}.  Despite the approximation, the resulting SCM is still defined in terms of $\theta$. In this manner the SCM retains the biological mechanisms and the interpretation of the Markov process model.

\noindent {\bf Create  ``observation" and ``intervention"  models} (Algorithm \ref{CF} lines \ref{obsModel}-\ref{metaprog}) In a probabilistic programming language, the deterministic functions in  \eqref{scm_model} are specified with a Dirac Delta distribution. However, at the time of writing, gradient-based inference in \emph{Pyro} produced errors when conditioning on a Dirac sample. We relaxed the Dirac Delta to allow a small amount of density.
\vspace{-4mm}
\begin{figure}[htbp!]
\begin{center}
\begin{tabular}{cc}
\includegraphics[width=0.43\columnwidth]{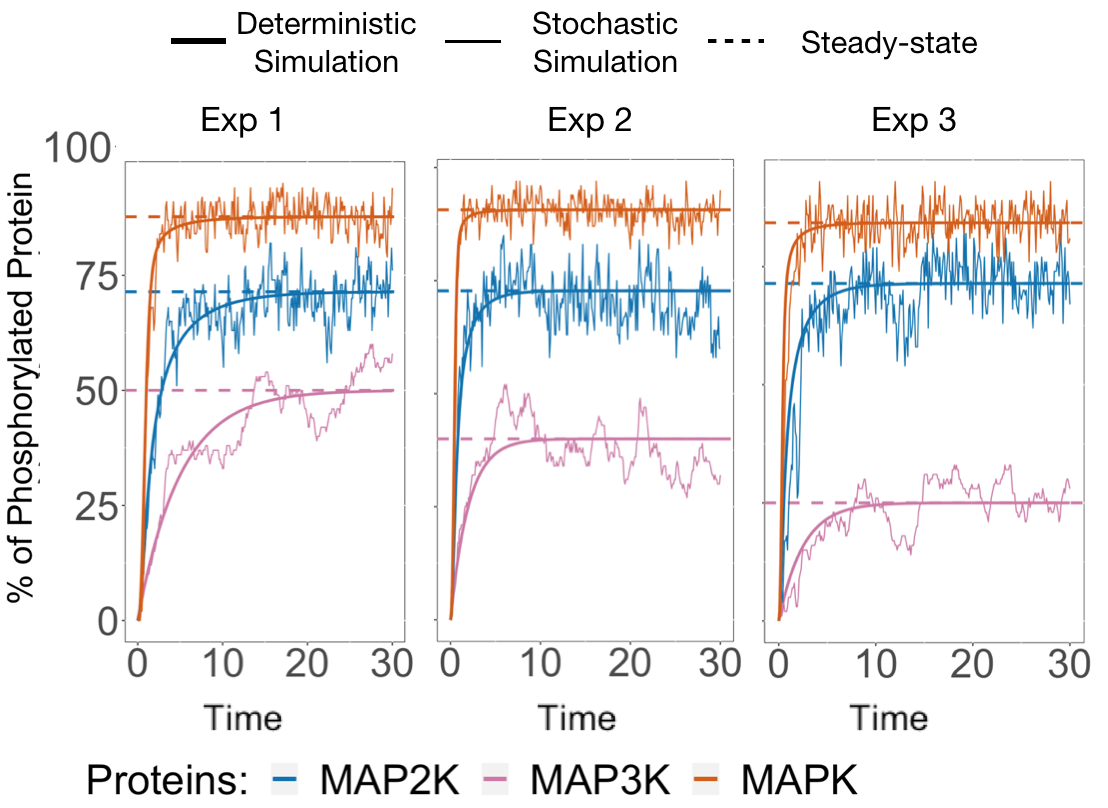} &
\includegraphics[width=0.53\textwidth]{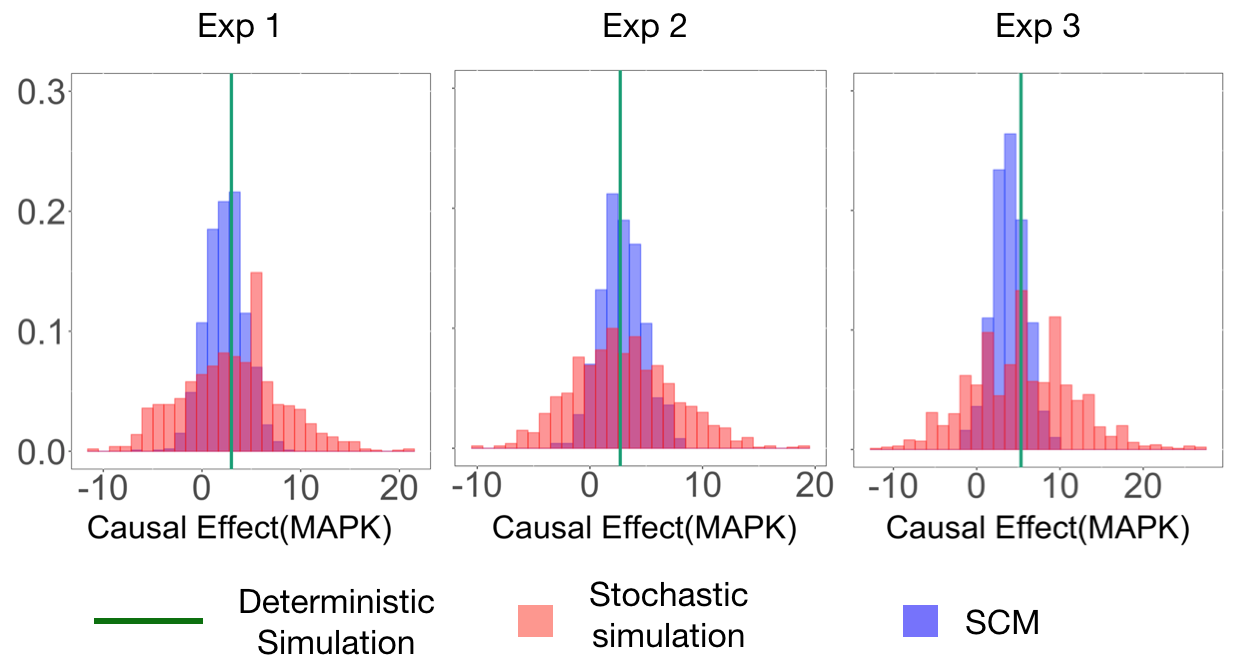} \\
(a) & (b)
\end{tabular}
\end{center}
\vspace{-.15in}
    \caption{\small \textbf{Case study 1 (MAPK)}. (a) Deterministic and stochastic trajectories of the active-state MAPK proteins. Horizontal lines are the expected values at equilibrium. (b) Histograms of causal effects, defined as differences between the ``observed" and the ``counterfactual" trajectories of $\text{MAP3K}$ at equilibrium.\label{fig:MapkSteadyState}}
 \vspace{-3mm}   
 \end{figure}

\noindent {\bf Infer noise distribution with observation model} (Algorithm \ref{CF} line \ref{update_noise})
We use stochastic variational inference (\cite{2012arXiv1206.7051H}) to infer and update $N_{\text{K3}}$, $N_{\text{K2}}$ and $N_{\text{K}}$ from the observation model, and independent Normal distributions as approximating distributions.

\noindent {\bf Simulate from intervention model with updated noise} (Algorithm \ref{CF} line \ref{simulate_cf})
After updating the noise distributions, we generate the target distribution of the intervention model.

\noindent {\bf Deterministic and stochastic counterfactual simulation and evaluation} (Algorithms 3 and 4 in Supplementary materials). \figref{fig:MapkSteadyState}(a) illustrates that the simulated trajectories converge in steady state. Since we rely on the Gaussian approximation to the Binomial in constructing $\mathbb{C}$, we would expect worse results if we were to set the rates  on or near the boundaries 0 and 100, where the approximation is weak.  \figref{fig:MapkSteadyState}(b) shows that for each experiment with different sets of rates, the causal effects from the SCM's counterfactual distribution are centered around the ground truth simulated deterministically using \eqref{3K_solution} and similar equations for \text{K2} and \text{K}. The SCM's distribution has less variance, likely due to the fact that ideal interventions in the SCM allow less variation than rate-manipulation-based interventions in the Markov process model.

\noindent {\bf{Evaluation under model misspecification}} \figref{fig:misspecification}(a) shows histograms from one of the 50 repetitions of the experiment conducted to evaluate the robustness of the SCM under model misspecification, and illustrates that the causal effects from the misspecified SCM is closer to true causal effect than the causal effect derived from a direct but misspecified simulation. Over the 50 repetitions, the absolute difference between the median of the true causal effect and the causal effect derived from the misspecified SCM is on average 0.343. The absolute difference between the median of the true causal effect and misspecified direct simulation is on average 1.03.

\vspace{-2mm}
\subsection{Case Study 2: The IGF signaling system}

\vspace{-2mm}

The derivations for the growth factor signaling system align closely with that of the motivating example and of the MAPK model.
For variable $X_i$ with parents $\mathbf{PA}_{\mathbb{M}, i}$, we partition each parent set into activators and inhibitors $\mathbf{PA}_{\mathbb{M}, i} = \{\mathbf{PA}^{\text{act}}_{\mathbb{M}, i},\mathbf{PA}^{\text{inh}}_{\mathbb{M}, i} \}$. The rate parameters are also partitioned into $\mathbf{v} = \{\mathbf{v}^{\text{act}}, \mathbf{v}^{\text{inh}}\}$. For each $X_i$ the probability for particle activation at equilibrium is:
\vspace{-2mm}
\begin{align}
\theta_{X_i}(\mathbf{PA}_{\mathbb{M}, i})= \frac{\mathbf{v}^{\text{act}}\mathbf{PA}^{\text{act}}_{\mathbb{M}, i}}{\mathbf{v}^{\text{act}}\mathbf{PA}^{\text{act}}_{\mathbb{M}, i} + \mathbf{v}^{\text{inh}}\mathbf{PA}^{\text{inh}}_{\mathbb{M}, i}}
\end{align}
Next, we derive an SCM using the same Normal approximation to the Binomial distribution as in the MAPK pathway.
\figref{fig:igf_ce}(b) plots deterministic and stochastic time courses for active states counts of the proteins in the pathway. \figref{fig:igf_ce}(c) illustrates that the counterfactual inference was successful despite the increased model complexity and size.

\begin{figure}[htbp!]
\begin{center}
\includegraphics[height=1.35in]{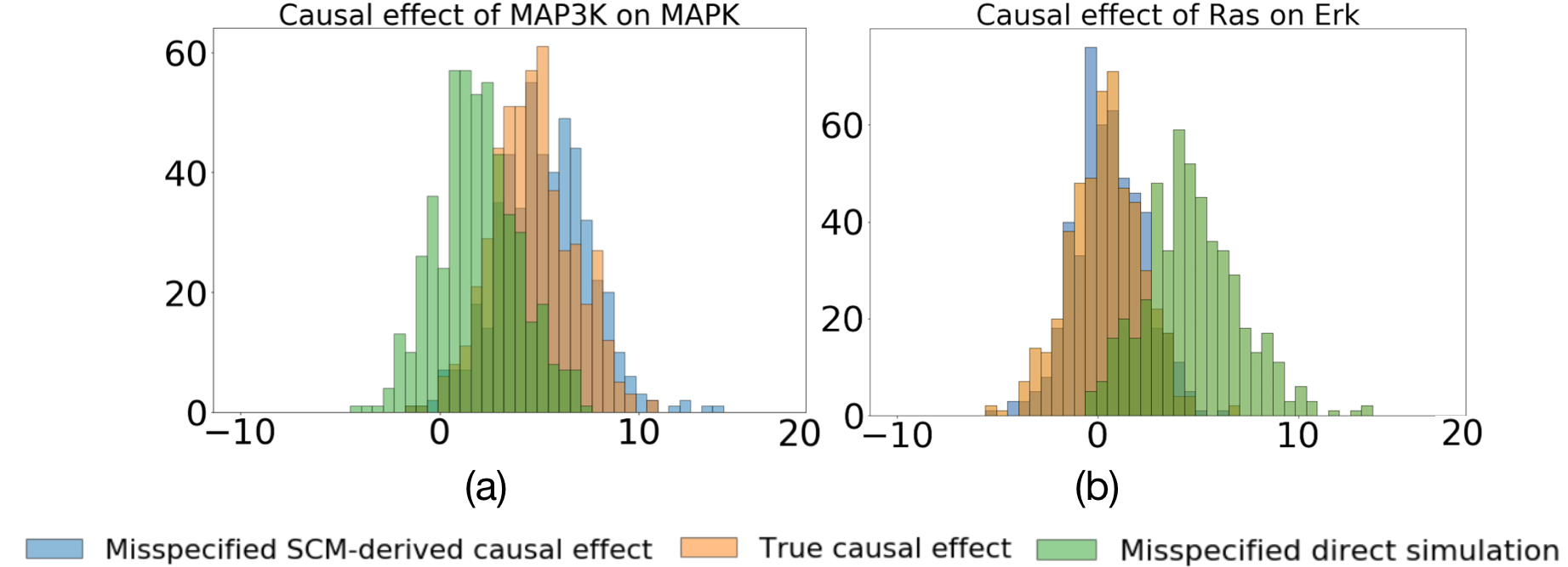}
\vspace{-.05in}
\caption{\small Histograms of causal effects, i.e. differences between the ``observed" and the ``counterfactual" trajectories at equilibrium, for one repetition of the evaluation. The causal effect from the misspecified SCM (blue histogram) is closer to true causal effect (orange histogram) than the causal effect derived from a direct but misspecified simulation (green histogram). (a) MAPK, (b) IGF.  \label{fig:misspecification}}
\end{center}
\end{figure}

\noindent {\bf{The evaluation under model misspecification}} Similarly to Case study 1, \figref{fig:misspecification}(b) illustrates that the causal effects from the misspecified SCM are closer to the true causal effect than the causal effect derived from a direct but misspecified simulation. Over the 50 repetitions, the absolute difference between the median of the true causal effect and the causal effect derived from the misspecified SCM is on average 7.563. The absolute difference between the median of the true causal effect and misspecified direct simulation is on average 92.55.


\vspace{-9mm}
\section{Discussion}
\vspace{-2mm}
This work proposed a practical approach for casting a Markov process model of a system at equilibrium as an SCM. Equilibrium counterfactual inferences using this SCM are anchored to the rate laws of the Markov process. We derived the specific steps of conducting counterfactual inference in real-life case studies of biochemical networks. The case studies illustrate that the counterfactual inference is consistent with the differences in the initial and the intervened upon trajectories of the Markov process, and makes the selection of interventions more robust to model misspecification. This approach opens many opportunities for future methodological research, such as extending this approach to models with cycles, a common feature of complex systems. Overall, this work is a step towards broader adoption of counterfactual inference in systems biology and other applications.

\clearpage
\bibliography{main}
\bibliographystyle{plain}

\end{document}


\vspace{-0.5cm}
\maketitle

%

\section{Supplementary data and algorithms}

\begin{table}[h!]
\centering
\scalebox{0.75}{
\begin{tabular}{c|cccccc}
& $v^{\text{act}}_{\text{K3}}$: MAP3K on & $v^{\text{inh}}_{\text{K3}}$: MAP3K off & $v^{\text{act}}_{\text{K2}}$: MAP2K on &  $v^{\text{inh}}_{\text{K2}}$: MAP2K off & $v^{\text{act}}_{\text{K}}$: MAPK on & $v^{\text{inh}}_{\text{K}}$: MAPK off  \\
\hline
Exp 1 & 0.1 & 0.1 & 0.1 & 2.0 & 0.1 & 1.0 \\
Exp 2 & 0.2 & 0.3 & 0.2 & 3.0 & 0.2 & 1.5 \\
Exp 3 & 0.1 & 0.3 & 0.5 & 5.0 & 0.3 & 4.0
\end{tabular}
}
\caption{\small The rates parameters in Case study 1 (MAPK).  Each row corresponds to a set $\mathbf{v}$.   \label{InitRates}}
\end{table}

\begin{table}[h!]
\centering
\scalebox{0.75}{
\begin{tabular}{c|ccccccccccc}
&
$v_{\text{SOS-EGFR}}^{\text{act}}$  &
$v_{\text{SOS-IGFR}}^{\text{act}}$  & 
$v_{\text{Ras-SOS}}^{\text{act}}$   & 
$v_{\text{PI3K-EGFR}}^{\text{act}}$ &
$v_{\text{PI3K-IGFR}}^{\text{act}}$ & 
$v_{\text{PI3K-Ras}}^{\text{act}}$  & 
$v_{\text{AKT-PI3K}}^{\text{act}}$  &
$v_{\text{Raf-Ras}}^{\text{act}}$   & 
$v_{\text{Raf-AKT}}^{\text{act}}$   &
$v_{\text{Mek-Raf}}^{\text{act}}$   &
$v_{\text{Erk-Mek}}^{\text{act}}$   \\
\hline
Rates & 0.01& 0.01& 0.01& 0.01& 0.01& 0.01& 0.01& 0.01& 0.01& 0.05& 0.05  
\end{tabular}
}
\caption{\small The activation rates parameters in Case study 2 (IGF). Activation rates are formatted as $v^{\text{act}}_{\text{child-parent}}$}
\end{table}

\begin{table}[h!]
\centering
\scalebox{0.75}{
\begin{tabular}{c|cccccccc}
&
$v_{\text{SOS}}^{\text{inh}}$      & 
$v_{\text{Ras}}^{\text{inh}}$      & 
$v_{\text{PI3K}}^{\text{inh}}$     & 
$v_{\text{AKT}}^{\text{inh}}$      &
$v_{\text{Raf}}^{\text{inh}}$ & 
$v_{\text{Raf-AKT}}^{\text{inh}}$  & 
$v_{\text{Mek}}^{\text{inh}}$      & 
$v_{\text{Erk}}^{\text{inh}}$ \\
\hline
Rates & 0.5& 0.5& 0.5& 0.5& 0.3& 0.01& 0.5& 0.5
\end{tabular}
}
\caption{\small The deactivation rates parameters in Case study 2 (IGF). All but one rate are auto-deactivation. $v^{\text{inh}}_{\text{RAF-ACT}}$ is formatted as $v^{\text{inh}}_{\text{child-parent}}$}
\end{table}

\medskip \noindent {\bf Runtime} All the experiments were performed on a Macbook pro with intel 2.2 GHz core i7 and 16 GB RAM. The counterfactual inference (Algorithm 2) on the MAPK and growth factor models with one sample was computed in 4.89 seconds and 10.4 seconds with stochastic variational inference. For the MAPK model, Algorithm \ref{ODEeval} with trajectories of length $30$ took $0.1$ seconds, and Algorithm \ref{StochSimEval} with $1000$ seeds took $1800$ seconds. For the IGF model, the Algorithm \ref{ODEeval} with trajectories of length $30$ took $0.9$ and Algorithm \ref{StochSimEval} with $300$ seeds took $1245$ seconds.

\clearpage
\setcounter{algorithm}{2}
\noindent
\begin{minipage}[t!]{.7\textwidth}
\raggedright

\begin{algorithm}[H]
\begin{small}
\caption{ \small \textit{Deterministic counterfactual simulation
\newline $~~~~~~~~~~~~~~~~~$ and evaluation with Markov process model}
\newline Inputs: $~~~$ Markov process model $\mathbb{M}$
\newline $~~~~~~~~~~~~~~~~~$ Rate sets $\mathbf{v}$ and $\mathbf{v}'$
\newline $~~~~~~~~~~~~~~~~~$ Equilibrium time point $T$
\newline $~~~~~~~~~~~~~~~~~$ Index of counterfactual intervention target $i$
\newline $~~~~~~~~~~~~~~~~~$ Index of counterfactual query target $j$
\newline $~~~~~~~~~~~~~~~~~$ Noise prior for SCM $\it{NPrior}$
\newline Output: $~~$ Histogram of causal effects
}
\label{ODEeval}
\begin{algorithmic}[1]
\Procedure{CF-DeterminSim}{$\mathbb{M}$, $\mathbf{v}$, $\mathbf{v}'$, $T$, $i$, $j$, $\text{NPrior}$}
\LineComment{{\footnotesize Simulate expected value at equilbrium (\eqref{limit} below) using $\mathbf{v}$}} 
\State $\mathbf{x} = E(\text{sim}(\mathbb{M}, \mathbf{v})[T]$) \label{ode_obs_sim}
\LineComment{{\footnotesize Simulate expected value at equilibrium (\eqref{limit} below) using $\mathbf{v}'$}}
\State $\mathbf{x}' = E(\text{sim}(\mathbb{M}, \mathbf{v}')[T]$) \label{ode_cf_sim}
\LineComment{{\footnotesize Calculate causal effects}}
\For {index $k$ in array $\Delta$}
\State $\delta_{\text{true}} = x'_j - x_j$
\LineComment{{\footnotesize Simulate CF value from SCM}}
\State $\mathbb{C} = \text{GetSCM}(\mathbb{M})$
\State $x_j^* \sim \text{CFQuery}(\mathbb{C}, \text{NPrior}, \mathbf{x},  x'_i)$ \label{scm_cf_sim_1}
\LineComment{{\footnotesize Calculate difference}}
\State $\Delta[k] = x^*_j - x_j$ \label{ode_ground_truth}
\EndFor
\LineComment{{\footnotesize Compare $\Delta$ to $\delta_{true}$}}
\State $\text{histogram}(\Delta, \text{verticle-line}=\delta_{\text{true}})$
\EndProcedure
\end{algorithmic}
\end{small}
\end{algorithm}

\end{minipage}
\newline
\begin{minipage}[t!]{.7\textwidth}
\raggedleft

\begin{algorithm}[H]
\begin{small}
\caption{ \small \textit{Stochastic counterfactual simulation
\newline $~~~~~~~~~~~~~~~~~$and evaluation with Markov process model}
\newline Inputs: $~~~$ Markov process model $\mathbb{M}$
\newline $~~~~~~~~~~~~~~~~~$ Rate sets $\mathbf{v}$ and $\mathbf{v}'$
\newline $~~~~~~~~~~~~~~~~~$ Equilibrium time point $T$
\newline $~~~~~~~~~~~~~~~~~$ Index of counterfactual intervention target $i$
\newline $~~~~~~~~~~~~~~~~~$ Index of counterfactual query target $j$
\newline $~~~~~~~~~~~~~~~~~$ List of random seeds $S$
\newline $~~~~~~~~~~~~~~~~~$ Noise prior for SCM \it{NPrior}
\newline Output: $~~$ Histogram of causal effects
}
\label{StochSimEval}
\begin{algorithmic}[1]
\Procedure{CF-StochSim}{$\mathbb{M}$, $\mathbf{v}$, $\mathbf{v}'$, $T$, $i$, $j$, $S$, $\text{NPrior}$}
\LineComment{{\footnotesize Simulate a deterministic equilibrium using $\mathbf{v}'$}}
\State $\mathbf{x}^{d'} = \text{sim}(\mathbb{M}, \mathbf{v}')[T]$ \label{cde_gt_cf_sim}
\For {index $k$ in S, \& collectors $\Delta_{\mathbb{M}}$, $\Delta_{\mathbb{C}}$}
\LineComment{{\footnotesize Simulate a stochastic equilbrium using $\mathbf{v}$}}
\State $\mathbf{x}^{s} = \text{sim}(\mathbb{M}, \mathbf{v}, \text{seed}=S[k])[T]$ \label{sde_obs_sim}
\LineComment{{\footnotesize Simulate a stochastic equilibrium using $\mathbf{v'}$}}
\State $\mathbf{x}^{s'} = \text{sim}(\mathbb{M}, \mathbf{v}', \text{seed}=S[k])[T]$ \label{sde_cf_sim}
\LineComment{{\footnotesize Calculate causal effects}}
\State $\Delta_{\mathbb{M}}[k] = x^{s'}_j - x^{s}_j$ \label{rand_causal_effect}
\LineComment{{\footnotesize Simulate CF value from SCM using $\mathbf{x}^{d'}$}}
\State $\mathbb{C} = \text{GetSCM}(\mathbb{M})$
\State $x_j^* \sim \text{CFQuery}(\mathbb{C}, \text{NPrior}, \mathbf{x},  x^{d'}_i)$ \label{scm_cf_sim_2}
\LineComment{{\footnotesize Calculate difference}}
\State $\Delta_{\mathbb{C}}[k] = x^*_j - x^{s}_j$ \label{scm_cf_sim_2b}
\EndFor
\LineComment{{\footnotesize Compare $\Delta_{\mathbb{C}}$ to $\Delta_{\mathbb{M}}$}}
\State $\text{overlayHistograms}(\Delta_{\mathbb{C}}, \Delta_{\mathbb{M}})$
\EndProcedure
\end{algorithmic}
\end{small}
\end{algorithm}

\end{minipage}

\clearpage
\section{Supplementary methods and proofs}

\subsection{Markov process equilibrium as causal Bayesian network}

In many domains, Markov process models can have arbitrary levels of granularity in terms of the components in the system and their interactions. This is certainly true in systems biology, where Markov process models (called stochastic kinetic models in this context) vary in details of chemical reactions between molecular species. In order to generalize the proposed procedure, we assume that the random variables in the Markov process model are {\it place invariants}.

\begin{definition}{\bf Place invariant}
A place invariant (also called p-invariant or s-invariant) is a set of model components with a constant sum across all the model states.  A minimal place invariant cannot be decomposed into smaller place invariants \cite{dubins1966invariant}.
\end{definition}

We assume a Markov process model with a unique equilibrium distribution and no cycles (in most cases, the cycles can be collapsed into a larger p-invariant).  Therefore, the equilibrium distribution is factorized according to a directed acyclic graph, given by solving each $\frac{dP(X(t))}{dt}$ for each variable $X$ in the model. We cast this distribution as a causal Bayesian network, with the conditional probability distributions given by the equilibrium distributions.

\subsection{Comparison to other intervention approaches}

In causal inference literature, a commonly used definition of intervention is the ``ideal" intervention, which {\it directly} sets the value of the target variable, and cuts off the influence of the targets' direct parents. This definition is adapted to dynamic processes in \cite{bongers2018random,dash2005restructuring,mooij2013ordinary}, where an ideal intervention fixes the target variable at a specific value throughout the transient states of the system until the equilibrium is reached, and then blocks all influence on the intervened variable with no side effects. The prior work on ideal interventions largely focused on the equivalence between the equilibrium outcome of the intervention on the dynamic model, and the ``do''-style ideal interventions on an equilibrium SCM. Reference \cite{dash2005restructuring} refers to this as the {\it equilibrium-manipulation commutability property}.

This manuscript uses a different approach. We define an intervention on a Markov process as a manipulation of the parameter rates, to {\it indirectly} achieve a desired equilibrium value of a target variable. Unless the manipulation sets the reaction rate to zero, the parent variables are still influencing the intervention target upon the intervention. This definition of the intervention is motivated by the biological application, where it is common to have directly manipulable rates (e.g., through a catalyst), and where the equilibrium values are typically an indirect result of the manipulation. 

In contrast to the prior work on ideal interventions in dynamic processes, this manuscript aims to cast an equilibrium probability model $\mathbb{G}$ as an SCM, in order to make useful counterfactual inference on that model.  We show that with zero or first order hazard functions, we can work directly with the solutions to a system of ordinary differential equations on the expectations of each variable in the system. These constitute what \cite{mooij2013ordinary} calls a {\it labeled set of equilibrium equations}. 

The distinction between manipulable and non-manipulable causes is treated in depth in the causal inference literature  \cite{pearl2018does, pearl2019interpretation}. Those interested in working with ideal interventions with our proposed approach could view the solutions to the expectation equations as a ``labeled set of equilibrium equations" in \cite{mooij2013ordinary}.  Since the expectation equations are deterministic, this can help investigate whether the equilibrium-manipulation commutability conditions in Lemma 1 of \cite{mooij2013ordinary} could apply to ideal interventions {\it in expectation} with our approach.

Those interested in working with ideal interventions with our proposed approach could view the solutions to the expectation equations as a ``labeled set of equilibrium equations" in \cite{mooij2013ordinary}.  Since the expectation equations are deterministic, this can help investigate whether the equilibrium-manipulation commutability conditions in Lemma 1 of \cite{mooij2013ordinary} could apply to ideal interventions {\it in expectation} with our approach.

\subsection{Details of the motivating example}

\subsubsection{Summary}
Assume that species $\mbox{Y}$ is regulated by species $\mbox{X}_1$ and $\mbox{X}_2$.  A particle of any of these three species is in either an active or inactive state.  $\mbox{X}_1$ is an activator of $\mbox{Y}$, meaning that an interaction with active $\mbox{X}_1$ converts inactive $\mbox{Y}$ into active $\mbox{Y}$.  $\mbox{X}_1$ is an deactivator. An interaction with active $\mbox{X}_2$ converts active $\mbox{Y}$ to inactive $\mbox{Y}$.  We can represent the reactions under a mass action kinetic assumption as follows:
\begin{eqnarray}
\mbox{X}_1^{\text{on}} + \mbox{Y}^{\text{off}}   &\overset{v_1}{\rightarrow} \ \mbox{X}_1^{\text{on}} + \mbox{Y}^{\text{on}} \  \ \mbox{and}\ \
\mbox{X}_2^{\text{on}} + \mbox{Y}^{\text{on}} &\overset{v_2}{\rightarrow} \ \mbox{X}_2^{\text{on}} + \mbox{Y}^{\text{off}} \label{e1}
\end{eqnarray}
$v_1$ and $v_2$ are the rate parameters for the two reactions.
In this example, $\mbox{X}_1$ is p-invariant, as the sum of active (``on") and inactive (``off") particles is constant.  The same holds for $\mbox{X}_2$ and $\mbox{Y}$.

\subsection{Building a probability model of the system}

Let $X_1(t)$, $X_2(t)$, and $Y(t)$ represent the total \emph{active-state} particle count in a cell of $\mbox{X}_1$, $\mbox{X}_2$, and $\mbox{Y}$ respectively at time $t$.  Let $T_y$ represent the total particle count (active and inactive) in a cell of $\mbox{Y}$, such that $T_y - Y(t)$ is the number of inactive particles of $\mbox{Y}$ at time $t$. 

Let $\pi(y, t) = P(Y(t) = y \mid T_y, X_1(t), X_2(t))$ represent the conditional probability distribution of $Y(t)$.  Each particle of $\mbox{Y}$ is in active state with some probability, i.e. a Bernoulli trial.  Therefore $Y(t)$ is the sum of Bernoulli random variables such that $\pi(y, t)$ is a Binomial distribution with $T_y$ trials.  

Let $\theta_{y}(X_1(t), X_2(t))$ denote the probability that a particle of $\mbox{Y}$ is in active state at time $t$.  This probability is needed to fully specify the Binomial distribution.  The following derivation demonstrates how $\theta_{y}(X_1(t), X_2(t))$ is a function of $X_1(t), X_2(t)$.  

\subsubsection{Finding the equilibrium probability distribution for $Y(t)$}

The \emph{hazard rate function} of a biochemical reaction is the probability that the reaction occurs in a given instant. It is determined by the particle counts at that instant, and by the rate parameters.  Let $h_1(Y(t))$ and $h_2(Y(t))$ represent the hazard rate functions for activation and deactivation respectively:
\begin{align}
h_1(Y(t)) &= v_1X_1(t) (T_y- Y(t)) \\
h_2(Y(t))  &= v_2X_2(t)Y(t) \label{eq:hazards}
\end{align}

Let $S_1$ and $S_2$ denote the change in particle count after reactions in  \eqref{e1}.
\begin{align}
S_1 &= 1 \nonumber \\
S_2  &=  -1 \nonumber
\end{align}

The Kolmogorov forward equations determine the change in $\pi(y, t)$ as the system evolves in time.  
\begin{align}
\frac{d}{dt}\pi(y, t) = \sum_{i = 1}^2 \left (h_i(y - S_i)\pi(y - S_i, t) - h_i(y)\pi(y, t) \right ) \nonumber
\end{align}

Let $E_{\pi}(.)$ denote the conditional expectation function over $\pi(y, t)$, i.e. $E_{\pi}(f(Y(t))) \equiv \sum_{y=0}^{T_y} f(y)\pi(y, t)$.

\medskip \begin{lemma} If the hazard functions are zero or first order, then solving for equilibrium of $\frac{d}{dt}E_{\pi}(Y(t))$ yields the equilibrium solution to the $\frac{d}{dt}\pi(y, t)$
\end{lemma}

\begin{proof}
By Kolmogorov's forward equation, the change in the expectation of $Y(t)$ is
\begin{align}
\frac{d}{dt}E_{\pi}(Y(t)) &= \frac{d}{dt}\sum_{y = 0} ^{T_y} y \pi(y, t) = \sum_{y=0}^{T_y} y \frac{d}{dt}\pi(y, t) \nonumber \\
&= \sum_{y = 0}^{T_y} y \left [ \sum_{i = 1}^2 \left (h_i(y - S_i)\pi(y - S_i, t) - h_i(y)\pi(y, t) \right )  \right ] \nonumber \\
&=  \sum_{i = 1}^2 \left ( \sum_{y = 0}^{T_y} y h_i(y - S_i )\pi(y - S_i \mid T_y, X_{1}(t), X_{2}(t)) -\sum_{y = 0}^{T_y} y  h_i(y)\pi(y, t)  \right ) \nonumber \\
&=   \sum_{i = 1}^2 \left (  \sum_{y = 0}^{T_y} (y + S_i ) h_i(y)\pi(y, t) -\sum_{y = 0}^{T_y} y  h_i(y)\pi(y, t) \right ) \nonumber \\
&=   \sum_{i = 1}^2 \left(  E_{\pi}((Y(t) + S_i ) h_i(Y(t))) - E_{\pi}(h_i(Y(t))) \right) \nonumber \\
&=   \sum_{i = 1}^2 S_i  E_{\pi} \left(h_i(Y(t))  \right) \nonumber
\end{align}

If hazards are zero or first order, then the linearity property of the expectation operator allows for an analytical solution.  Without loss of generality, we demonstrate this with the motivating example. Substituting in the hazard functions in \eqref{eq:hazards}:

\begin{align}
\frac{d}{dt}E_{\pi}(Y(t)) &=   \sum_{i = 1}^2 S_i  E_{\pi} \left( h_i(Y(t))\right) \nonumber \\
&=  E_{\pi} v_1X_1(t) (T_y- Y(t)) - E_{\pi} v_2X_2(t)Y(t) \nonumber \\
&=   v_1X_1(t)T_y - v_1X_1(t) E_{\pi}(Y(t)) - v_2X_2(t) E_{\pi}(Y(t)) \nonumber \\
&=   v_1X_1(t)T_y - \left(v_1X_1(t) + v_2X_2(t)\right)  E_{\pi}(Y(t)) \label{dE}
\end{align}	

Let $\theta_{y}(X_1(t), X_2(t))$ denote the probability that a particle of $\mbox{Y}$ is in active state at time $t$.  $\theta_{y}(X_1(t), X_2(t))$ is a function $X_1(t), X_2(t)$.  This follows from the fact that $\pi(y, t)$ has a Binomial distribution, and therefore $\theta_{y}$ is determined by the expectation $E_{\pi}(Y(t))$.
\begin{align}
\theta_{y}(X_1(t), X_2(t)) \equiv \frac{E_{\pi}(Y(t))}{T_y} \label{param}
\end{align}
Substituting Equation \ref{param} into \ref{dE}:
\begin{align}
\frac{d}{dt}E_{\pi}(Y(t)) &= v_1X_1(t)T_y - \left(v_1X_1(t) + v_2X_2(t)\right)  T_y \theta_{y}(X_1(t), X_2(t)) \nonumber \\
\frac{d}{dt}\theta_{y}(X_1(t), X_2(t)) &= v_1 X_1(t) - \left(v_1X_1(t) + v_2X_2(t) \right)  \theta_{y}(X_1(t), X_2(t)) \label{d_delta}
\end{align}

The analytical solution is:

\begin{align}
\theta_{y}(X_{1}(t), X_{2}(t)) &= e^{-t(v_1X_{1}(t) + v_2X_{2}(t))} + \frac{v_1X_{1}(t)}{v_1X_{1}(t) + v_2X_{2}(t)}
\label{dyn_theta1}
\end{align}

Therefore the probability distribution of $Y(t)$ is given by 
\begin{align}
Y(t) \sim \text{Binomial}(T_y, \theta_{y}(X_{1}(t), X_{2}(t))) \label{dyn_prob}
\end{align} 

$\theta_{y}(X_{1}(t), X_{2}(t))$ achieves \emph{steady-state} when $\frac{d}{dt}\theta_{y}(X_1(t), X_2(t)) = 0$.  The solution is found by setting the left-hand side in \eqref{d_delta} to 0, or alternatively, taking the limit in time of \eqref{dyn_theta1}
\begin{align}
\underset{t \to \infty}{lim} \theta_{y}(X_{1}(t), X_{2}(t)) 
&= \frac{v_1 X_1(t)}{v_1 X_1(t) + v_2X_2(t)}\label{limit}
\end{align}
$\theta_{y}(X_1(t), X_2(t))$ is the only component of the Binomial probability distribution of $Y(t)$ that varies in time. The steady-state solution of \eqref{dyn_theta1} also provides the steady-state distribution  of $Y(t)$ (also referred to as stationary or invariant distribution in stochastic process literature).  For simplicity we assume that the counts of active $\mbox{X}_1$ and $\mbox{X}_2$ are also the results of processes with stationary distributions.  Let $Y$, $X_1$ and $X_2$ represent steady state active particle counts for $\mbox{Y}$, $\mbox{X}_1$, and $\mbox{X}_2$.   Then the steady-state distribution of $Y$ is given by:
\begin{align}
\theta_{y}(X_{1}, X_{2}) = \frac{v_1 X_1}{v_1 X_1 + v_2X_2} \label{steady_theta} \\
Y \sim \text{Binomial} \left (T_y, \theta_{y}(X_{1}, X_{2}) \right) \label{stat_prob}
\end{align}
\end{proof}

\subsection{Connections to causal constraint models}
The rate laws in a dynamic model describe the mechanistic relationships between variables of the model. Our goal is to build a causal model of the system at equilibrium that is faithful to these relationships.

However, particular sets of interventions could mutate the mechanisms underpinning these relationships. For example, the mechanisms in the dynamic model may give rise to a conservation law, i.e., that the sum of the values of a particular subset of variables is constant in time, though the values themselves may vary. In would be possible to specify interventions that violate this conservation law. The equilibrium SCM model assumes that the mechanisms connected to this conservation law are invariant and has no way of prohibiting a set of interventions that violate this assumption. The SCM's predictions of the equilibrium behavior of a set of interventions that violate this conservation law would be inconsistent with simulated interventions from the dynamic model.

Blom et al. address this problem by introducing causal constraint models (CCMs), an extension to SCM models of the equilibrium of dynamic models \cite{blombeyond}. CCMs explicitly identify sets of interventions for which the functional relationships derived from the underlying dynamic model are invariant.

Our work illustrates the process of converting a dynamic model to an equilibrium SCM and demonstrates counterfactual inferences consistent with the dynamic model ground truth. We apply simplifying assumptions to our dynamic models designed to avoid the conflicts between SCM intervention and the dynamic model mechanism that Blom et al.'s work addresses. In the following subsections, we show that the examples we use in this work do not have any intervention constraints that would warrant the use of a CCM.

The case studies used in this work are special cases of the enzyme kinetic reactions that take the form:

\begin{align}
E + S &\underset{K_{\text{off}}}{\overset{K_{\text{on}}}{\rightleftharpoons}}[ES] \overset{K_2}{\rightarrow} E + P \nonumber \\
P &\overset{v_{\text{off}}}{\rightarrow}S \nonumber 
\end{align}

where $E$ is a enzyme, $S$ is a substrate, and $P$ is a product. Blom et al. in constrast use the following enzyme model as an example:

\begin{align}
\varnothing &\overset{K_0}{\rightarrow} S \nonumber \\
E + S &\underset{K_{\text{-1}}}{\overset{K_{\text{1}}}{\rightleftharpoons}}[ES] \overset{K_2}{\rightarrow} E + P \nonumber \\
P &\overset{K_3}{\rightarrow} \varnothing  \nonumber 
\end{align}

Our enzyme model is a special case of the Blom et al. model where substrate does not appear from nothing nor substrate disappear into nothing, but rather product converts back to product.

Further, we simplify the model such that it collapses over the intermediate compound (simplifying from Michaelis–Menten kinetics to mass action kinetics): 

\begin{align}
E + S & \overset{v_{\text{on}}}{\rightarrow} E + P \nonumber \\
P &\overset{v_{\text{off}}}{\rightarrow}S \nonumber 
\end{align}

Finally, when modeling the hazard rates in our Markov process model, we incorporate the fact that in our model there is a conservation law between product and substrate.  Let $P(t)$ and $S(t)$ be the total amount of product and substrate at time $t$, and $T = P(t) + S(t)$ be the unchanging total. Instead of modeling the production rate law of $P$ as $v_{\text{on}}E(t)S(t)$, we eliminate the variable $S(t)$ from our model and use $v_{\text{on}}E(t) (T- P(t))$.  This, combined with the elimination of the compound $[ES]$ from the model, remove the need for the causal constraints outlined in Blom et al. enzyme model. 

We believe that our work and Blom et al.'s CCM framework are complimentary. CCM's could be used to avoid making simplifying assumptions when they are not appropriate relative to the complexity of the dynamic system. Our approach of modeling the dynamic model as a Markov process in order to derive a probability model of equilibrium, then finding an SCM that entails that probability model, is unexplored in CCMs. We believe this would be a fruitful avenue for future work.

\subsection{Proof the inverse Binomial CDF transform is a monotonic conversion.}

\theoremstyle{definition}
\begin{definition}{\bf Monotonic condition.}
A variable $Y$ is said to be monotonic relative to variable $X$ in a given structural causal model if and only if, given $X = x$ and noise variable $N = n$, the structural assignment $f_Y(x, n)$ is monotonic in $x$ for all $n$. If the monotonicity condition is true, then $E(Y|do(X=x)) >= E(Y|do(X=x')) \Rightarrow f_Y(x, n) \geq f_Y(x', n)\  \forall n$ \cite{oberst2019counterfactual,pearl2009causal}.
\end{definition}

\theoremstyle{definition}
\begin{definition}{\bf Monotonic conversion.}
A monotonic conversion is a conversion of a probabilistic generative model of $Y$ to a structural assignment (which assigns a value of $Y$ to a deterministic function of a random noise input) such that the assignment satisfies the monotonic condition.
\end{definition}

\begin{lemma}
Let $N$ be a noise variable with a $\text{Uniform}(0, 1)$, and let $n$ be a sample of $N$.  Let random variable $Y$ be generated from a probabilistic generative model $Y\sim\text{Binomial}(T, \theta(x))$.  Let $T$ be the total number of trials, and $\theta(x)$ be the success probability, where $\{x, \theta(x) : x \in \mathbb{N}, 0 < \theta(x) < 1\}$. Assume that $\theta(x)$ is monotonic in $x$ (as in \eqref{steady_theta}, except with $X_2$ held constant). Let $F^{-1}(\theta(x), T, n)$ denote the inverse Binomial CDF of $Y$, parameterized by $\theta(x)$ and $T$.  If $E(Y|do(X=x)) \geq E(Y|do(X=x'))$, then the inverse CDF of $Y$ $f_Y(x, n) = F^{-1}((\theta(x), T, n)$ is a monotonic conversion.  
\end{lemma}

\begin{proof}

Let $y = E(Y|do(X=x))$ and $y' = E(Y|do(X=x'))$. Given $y$ and $y'$, there exists a value of $n^* \in (0,1)$ such that $y = F^{-1}(\theta(x), T, n^*)$ and $y' = F^{-1}(\theta(x'), T,  n^*)$.  Therefore, if $y \geq y'$ then $F^{-1}(\theta(x), T, n^*) \geq F^{-1}(\theta(x'), T, n^*)$.

\end{proof}

\subsection{Poisson distribution example}

The examples and case studies in the manuscript each had equilibrium distributions that factored into binomial distributions.  However, these results are not specific to the Binomial distribution.  The following illustrates a similar model that works with the Poisson distribution.
Suppose that $T_y$ in Eq. \ref{stat_prob} were unknown. Within the probabilistic modeling framework, we can model $T_y$ as a latent variable with distribution $\pi_T$.    

\begin{align}
T_y &\sim \pi_T \\
\theta_{y}(X_{1}, X_{2}) &= \frac{v_1 X_1}{v_1 X_1 + v_2X_2} \label{steady_theta} \\
Y &\sim \text{Binomial} \left (T_y, \theta_{y}(X_{1}, X_{2}) \right) \label{stat_prob_poisson_0}
\end{align}

A useful result from hierarchical Bayesian modeling is to set $\pi_T$ to a Poisson distribution with parameter $\lambda$, which simplifies the model as follows \cite{raftery1988inference}:

\begin{align}
\theta_{y}(X_{1}, X_{2}) &= \frac{v_1 X_1}{v_1 X_1 + v_2X_2} \label{steady_theta} \\
Y &\sim \text{Poisson} \left (\lambda \theta_{y}(X_{1}, X_{2}) \right) \label{stat_prob_poisson_1}
\end{align}

We can then set the structural assignment for $Y$ in the SCM using an inverse Poisson CDF. Let $F^{-1}_{\text{Pois}}(\mu, n)$ be the inverse CDF transform that given a parameter $\mu$ and a variable $n$ sampled from a uniform on the unit interval, returns a sample of a Poisson-distributed random variable distributed according to a Poisson distribution with mean $\mu$:

\begin{align}
n &\sim \text{Uniform}(0, 1) \\
Y &\sim F^{-1}_{\text{Pois}}(\lambda \theta_{y}(X_{1}, X_{2}), n) \label{stat_prob_poisson_2}
\end{align}

In general, the approach outlined in this manuscript works with any closed-form equilibrium conditional distributions derived from the underlying Markov process model.

\bibliography{main}
\bibliographystyle{plain}